\documentclass[letterpaper, 10 pt, conference]{ieeeconf}  

\IEEEoverridecommandlockouts                              

\usepackage{amsmath}
\DeclareMathOperator*{\argmin}{arg\,min}
\usepackage{cite}
\usepackage{amsmath,amssymb,amsfonts}
\usepackage{graphicx}
\usepackage{textcomp}
\usepackage{xcolor}
\usepackage{booktabs}
\usepackage[ruled,linesnumbered, vlined]{algorithm2e}
\DontPrintSemicolon
\SetCommentSty{myCommentStyle}

\usepackage{amsfonts}
\let\labelindent\relax
\usepackage{enumitem}
\usepackage{amsmath}
\usepackage{graphicx}
\usepackage{subfig}
\usepackage[margin=0.792in]{geometry}
\usepackage{bbm}
\usepackage{algpseudocode} 
\usepackage{multirow}
\usepackage{xcolor}

\makeatletter
\DeclareRobustCommand\onedot{\futurelet\@let@token\@onedot}
\def\@onedot{\ifx\@let@token.\else.\null\fi\xspace}
 
\def\ie{\emph{i.e}\onedot} 
\def\cf{\emph{c.f}\onedot}

\makeatother

\def\BibTeX{{\rm B\kern-.05em{\sc i\kern-.025em b}\kern-.08em
    T\kern-.1667em\lower.7ex\hbox{E}\kern-.125emX}}

\definecolor{darkgreen}{RGB}{0, 100, 0}   
\definecolor{DarkMagenta}{rgb}{0.7, 0.0, 0.7}
\newcommand{\mznote}[1]{\textcolor{green}{\textbf{MZ: #1}}}

\title{\LARGE \bf
VARP: A Visual Adaptive Rearrangement Planner for Non-Monotone Object Rearrangement Planning
}

\author{Hanwen Ren, Meng Zheng, Zhongpai Gao, Anwesa Choudhuri,\\ Benjamin Planche, Terrence Chen and Ziyan Wu 
}

\begin{document}

\thispagestyle{empty}
\pagestyle{empty}


\twocolumn[{
\begin{@twocolumnfalse}
\maketitle
\begin{center}
    \captionsetup{type=figure}
    \includegraphics[trim={0.6cm 0.2cm 0.6cm 0.2cm},clip,width=\textwidth]{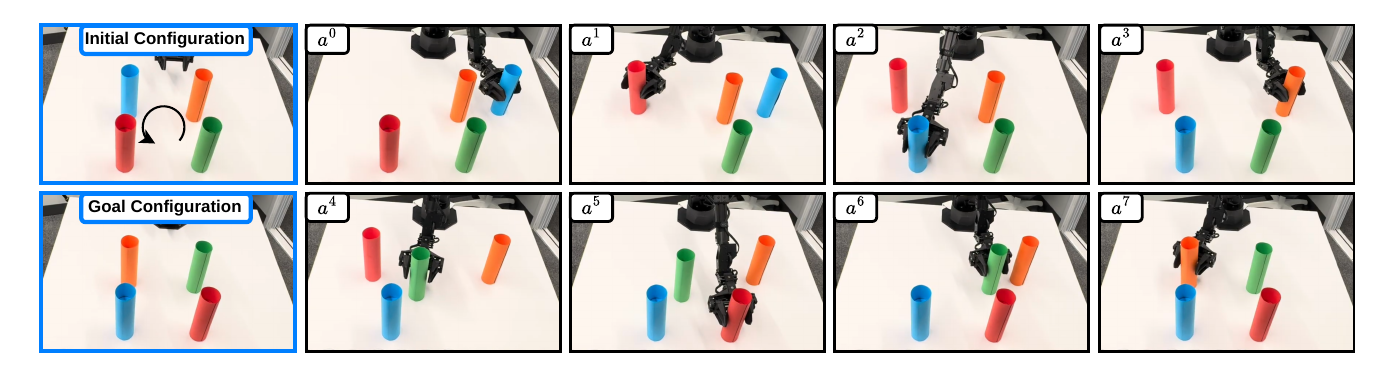}
    \captionof{figure}{Real-world execution of VARP on a challenging non-monotone four-object rearrangement task requiring an in-place 90-degree rotation. Our method continuously updates and predicts the next pick-and-place action based on the current and goal configurations until the goal arrangement is achieved, requiring only 0.118 s of planning time for the entire task.}
    \label{fig:real_exp1}%
\end{center}
\end{@twocolumnfalse}
}]

\begin{abstract}


Object rearrangement planning in cluttered environments is often non-monotone, requiring objects to be temporarily relocated before reaching their final destinations. Although search-based planners can solve such tasks, they incur substantial computational cost and require expensive replanning when the environment changes. We propose the \textbf{V}isual \textbf{A}daptive \textbf{R}earrangement \textbf{P}lanner (VARP), which learns non-monotone rearrangement directly from visual state representations generated from expert demonstrations. VARP predicts pick-and-place actions from the current and goal visual state representations, and incorporates an Action Refinement Strategy (ARS) to improve action feasibility and a Progress Enforcement Strategy (PES) to mitigate progress stagnation. By continuously updating actions from the latest visual state representations, VARP adapts online to disturbances without replanning from scratch or performing online combinatorial search. In challenging rearrangement benchmarks, VARP achieves a 98.3\% success rate with only 0.4 s planning time per task, 97.5\% success under execution-time disturbances, and 94.5\% under continuous severe disturbances, while generalizing effectively to unseen object counts. These results demonstrate that VARP provides a scalable and robust solution for non-monotone object rearrangement.

\end{abstract}

\section{INTRODUCTION}

Object rearrangement planning is a fundamental capability for robotic manipulation, with applications in warehouse automation, household assistance, manufacturing, and logistics \cite{han2017high}. Given an initial object arrangement and a desired target configuration, the robot must determine a sequence of prehensile pick-and-place actions that transform the scene while avoiding collisions and satisfying task constraints \cite{pan2022algorithms}. In cluttered environments, achieving the target arrangement often requires non-monotone planning, where objects must be temporarily relocated to an intermediate buffer region before they can be placed at their final destinations. Such non-monotone rearrangement problems are known to be NP-hard in general \cite{wilfong1988motion}, making the task particularly challenging as scene complexity increases.

Classical rearrangement planners typically formulate the problem as a search over object relocation sequences and rely on task-and-motion planning, sampling-based methods \cite{king2016rearrangement}, or tree search algorithms \cite{goyal2022ifor} to identify feasible solutions. While these approaches can produce high-quality plans, they often incur substantial computational cost due to repeated collision checking, dependency reasoning, and motion planning. Furthermore, when the environment changes during execution, planners frequently require expensive replanning procedures to recover from disturbances, limiting their applicability in dynamic settings.

Recent advances in imitation learning and vision-based manipulation have motivated the use of learning-based policies for rearrangement tasks\cite{yuan2018rearrangement}. By learning an observation-to-action mapping directly from expert demonstrations, neural policies can amortize planning effort and generate actions with significantly lower inference latency \cite{qureshi2021nerp}. However, CNN-based or Transformer-based policies approximate the observation-to-action mapping from finite data and may produce errors under perceptual uncertainty or distribution shift. Unlike search-based planners which explicitly reason about state transitions and action feasibility, existing learning-based approaches often struggle in cluttered non-monotone environments, where small prediction errors may lead to infeasible actions, self-relocations, or repetitive behaviors that prevent task completion \cite{tang2023selective}. Achieving the efficiency of learned policies while retaining the robustness of classical planning thus remains an open challenge.

To this end, we propose a \textbf{V}isual \textbf{A}daptive \textbf{R}earrangement \textbf{P}lanner (\textbf{VARP}) for non-monotone object rearrangement planning. VARP combines a goal-conditioned visual policy that efficiently predicts rearrangement actions from visual state representations with two lightweight inference-time reasoning mechanisms that improve action feasibility and long-horizon execution reliability. Executed in a closed-loop manner, VARP continuously updates its decisions based on the latest scene configuration, enabling robust adaptation to execution-time disturbances without expensive online replanning or combinatorial search. Extensive experiments demonstrate that VARP achieves superior success rates and substantially lower planning latency than state-of-the-art search-based planners while exhibiting strong robustness to environmental disturbances and generalization to out-of-distribution scenarios. Overall, the main contributions of this work are summarized as follows:

\begin{itemize}
\item We propose VARP, a visual adaptive rearrangement planner that integrates a learning-based goal-conditioned visual policy with lightweight inference-time reasoning for efficient and reliable non-monotone object rearrangement planning.

\item We introduce two lightweight inference-time reasoning mechanisms: the Action Refinement Strategy (ARS), which performs diversity-aware candidate exploration and feasibility validation, and the Progress Enforcement Strategy (PES), which performs short-horizon rollout reasoning to mitigate progress stagnation without expensive replanning.

\item We conduct extensive evaluations on standard rearrangement, disturbance recovery, out-of-distribution, and real-world scenarios, demonstrating that VARP achieves high success rates, short planning time, and robust performance across diverse evaluation settings.
\end{itemize}

\section{RELATED WORK}

\textbf{Object rearrangement planning:}
Object rearrangement planning has been widely studied in robotics, where the goal is to move multiple objects from an initial configuration to a target configuration while avoiding collisions and respecting robot motion constraints~\cite{stilman2008planning,  ren2024multi,krontiris2015dealing,king2016rearrangement, ren2026multi }. Classical approaches typically formulate rearrangement as a combinatorial search problem and rely on task-and-motion planning, sampling-based planning, or tree search to reason over object dependencies and intermediate placements~\cite{labbe2020monte, chen2024language}. These methods can provide strong planning performance in cluttered scenes; however, their computational cost often increases rapidly with the number of objects and the complexity of the required non-monotone reasoning \cite{wang2022efficient,kim2025integrating}. In contrast, VARP learns a goal-conditioned visual policy from expert demonstrations and performs fast closed-loop inference without solving a new combinatorial search problem from scratch at test time.

\textbf{Learning-based rearrangement policies:}
Recent learning-based approaches have investigated neural network-based policies that predict pick-and-place actions directly from visual observations~\cite{zeng2021transporter,shridhar2021cliport, ren2024neural}. Other works further incorporate language, spatial relations, or multimodal prompts for goal-conditioned and semantic rearrangement~\cite{liu2022structformer,shridhar2022perceiveractor,jiang2023vima}. These approaches reduce online planning cost by amortizing decision making into a learned policy, but they often struggle in cluttered non-monotone settings where intermediate relocations are required and small action errors can lead to invalid or unproductive behaviors. Unlike prior learning-based approaches that rely solely on feed-forward policy inference, VARP tightly couples a learned goal-conditioned visual policy with lightweight inference-time reasoning, enabling adaptive closed-loop planning without combinatorial search or expensive replanning.


\textbf{Closed-loop and reactive rearrangement:}
Closed-loop rearrangement is important as execution may deviate from the planned trajectory due to perception errors, actuation uncertainty, or external disturbances \cite{kappler2018real}. Prior work has explored reactive manipulation through online replanning, task-and-motion replanning, and model-predictive planning frameworks that repeatedly recompute actions based on updated observations \cite{liu2021deep,bowen2014closed}. While these approaches can improve execution robustness, their computational cost can become substantial in cluttered rearrangement settings. In contrast, VARP uses visual feedback at every step and directly predicts the next pick-and-place action from the current and goal visual state representations.

\section{PROBLEM FORMULATION}
We consider a non-monotone object rearrangement planning problem in a confined workspace $\mathcal{W} \subset \mathbb{R}^3$ containing $n$  geometrically identical but uniquely identifiable movable cylindrical objects $\mathcal{O} = \{o_1,\dots, o_n\}$. 
The workspace is modeled as a cabinet-like enclosure with a single front opening and enclosed boundaries on all other sides. Its inner surface dimensions along the $X$ and $Y$ axes are denoted by $(d_x, d_y)$, respectively.
Inside the workspace, each object $i\in \{1,...,n\}$ occupies a collision-free placement location $p_{o_i} = (x_{o_i}, y_{o_i})$, where $p_{o_i} \in \mathcal{P} \subset \mathbb{R}^2$. At any time step $t$, the placement locations of all objects define an object placement configuration $s^t = \{p_{o_1}^t,\dots, p_{o_n}^t\} \in \mathcal{S}$, with $\mathcal{S}$ denoting the configuration space and $s^t[o_i] = p_{o_i}^t$ indicating the object $o_i$ occupies location $p_{o_i}^t$. Finally, the volume occupied by all objects at time $t$ is written as $V(s^t) \subset \mathbb{R}^3$. \par
A robot manipulator $\mathcal{R}$ equipped with a parallel-jaw gripper is placed in front of the workspace opening at location $p_m$ in order to perform prehensile object relocation actions.
A rearrangement action at time step $t$ is defined as a composite pick-and-place action:
\begin{equation}
a^t = \left(a_{\mathrm{pk}}^t, a_{\mathrm{pl}}^t\right)
\label{eq:at_def}
\end{equation}
where $a_{\mathrm{pk}}^t$ selects and grasps an object $o_i$ from its current placement $s^t[o_i]$, and $a_{\mathrm{pl}}^t$ relocates the grasped object to a target placement $s^{t+1}[o_i]$. 
The action induces the state transition:
\begin{equation}
    s^{t+1} = f(s^t, a^t)
\label{eq:state_transition}
\end{equation}
The $d$-degree-of-freedom robot configuration space is denoted by $\mathcal{Q}\subset\mathbb{R}^d$. During a rearrangement action $a^t$ manipulating object $o_i$, the robot follows a motion trajectory $\tau(a^t)=\{q_0^t,\ldots,q_k^t\}$, where each $q_j^t\in\mathcal{Q}$ denotes an intermediate robot configuration. The corresponding swept volume of the robot is denoted by $V(\tau(a^t))\subset\mathbb{R}^3$. The action $a^t$ is considered feasible under the current configuration $s^t$ if the corresponding robot trajectory is collision-free with respect to all non-manipulated objects:
\begin{equation}
V(\tau(a^t))
\;\cap\;
V\!\left(\left\{\,s^t[o_j]\;\middle|\;o_j\in\mathcal{O},\,o_j\neq o_i\right\}\right)
=\emptyset 
\label{eq:valid_action}
\end{equation}
We formulate the object rearrangement task as a deterministic goal-conditioned planning problem defined by the tuple
\begin{equation}
    \mathcal{P}=(\mathcal{S},\mathcal{A},f,s^0,s^g)
\label{eq:planning}
\end{equation}
where $\mathcal{S}$ denotes the object configuration space, $\mathcal{A}$ denotes the space of feasible pick-and-place actions, $f:\mathcal{S}\times\mathcal{A}\rightarrow\mathcal{S}$ is the deterministic state transition function, and $s^0$ and $s^g$ denote the initial and goal configurations, respectively. The objective is to develop a planner that sequentially selects actions $a^t\in\mathcal{A}$ to transform the current configuration $s^t$ toward the goal configuration $s^g$. The resulting action sequence $\mathcal{T}=\{a^0,\ldots,a^T\}$ constitutes a feasible rearrangement plan that transforms the workspace from $s^0$ to $s^g$.

\begin{figure*}[t]
    \centering
    \includegraphics[width=\textwidth]{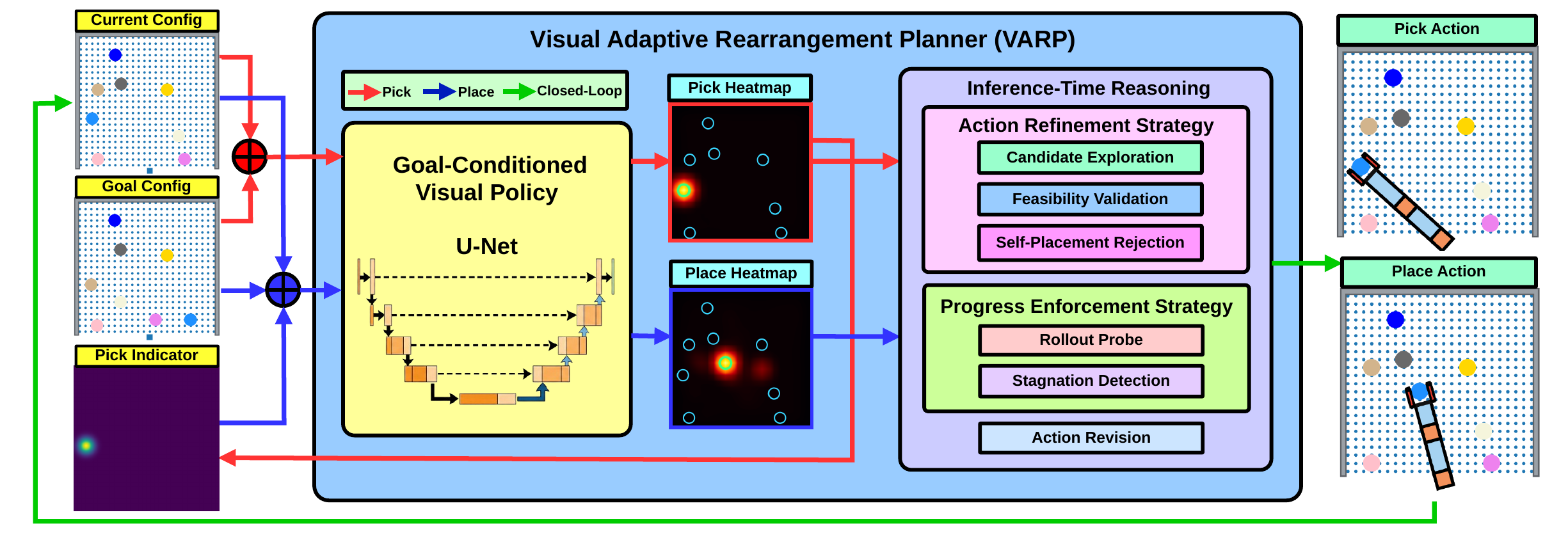}
    \caption{Architecture of the proposed Visual Adaptive Rearrangement Planner (VARP). A goal-conditioned visual policy first predicts pick and place heatmaps, which are subsequently refined by the Action Refinement Strategy (ARS) and Progress Enforcement Strategy (PES) into executable rearrangement actions through lightweight inference-time reasoning.}
    \label{fig:pipeline}
    \vspace{-0.5cm}
\end{figure*}

\section{METHOD}
\subsection{Method Overview}

To solve this problem, we propose the \textbf{V}isual \textbf{A}daptive \textbf{R}earrangement \textbf{P}lanner (VARP), a goal-conditioned closed-loop planner that generates one rearrangement action at a time using visual feedback from the environment. VARP consists of a learned goal-conditioned visual policy augmented with two inference-time reasoning modules: the \textbf{A}ction \textbf{R}efinement \textbf{S}trategy (ARS) and the \textbf{P}rogress \textbf{E}nforcement \textbf{S}trategy (PES), as illustrated in Fig.~\ref{fig:pipeline}. Given visual cues rendered from the current state $s^t$ and goal state $s^g$, the visual policy first outputs pick and place heatmaps. ARS then transforms the raw policy predictions into feasible rearrangement actions by promoting action diversity, enforcing motion feasibility, and rejecting ineffective self-placement actions. PES further evaluates the selected action through a short-horizon rollout probe and revises it when progress stagnation is detected. The resulting action is executed, producing a new state $s^{t+1}$, and the process repeats until the goal configuration is reached.

\subsection{Goal-Conditioned Visual Policy}

Rather than solving the rearrangement problem through explicit symbolic search, we learn a goal-conditioned visual policy that predicts the next rearrangement action directly from visual representations. The underlying state is the object placement configuration $s^t$, while the planner receives a rendered visual state representation of the configuration. We assume the current object configuration is provided by an external perception or object-tracking system. VARP operates on rendered visual state representations generated from these estimated object configurations rather than directly on raw RGB images. Specifically, given the current object configuration $s^t$ and the goal configuration $s^g$, we obtain their corresponding visual representations through a rendering function $\Phi(\cdot)$, :
\begin{equation}
I^t = \Phi(s^t), \qquad I^g = \Phi(s^g)
\end{equation}
where $I^t$ and $I^g$ denote the current and goal visual state representations, respectively. The rendering function $\Phi(\cdot)$ converts the object configuration into a fixed-resolution visual state representation while preserving the spatial relationships among objects. Geometrically identical objects are rendered using unique fixed colors, allowing the visual policy to distinguish individual objects. 

Then, the learned visual policy is defined as
\begin{equation}
\hat{a}^t = \pi_\theta(I^t, I^g)
\end{equation}
where $\pi_\theta$ is a visual policy model by a neural network with learnable parameters $\theta$, and $\hat{a}^t$ denotes the pick-and-place action predicted by the planner, consisting of a pick action $\hat{a}
_{\mathrm{pk}}^t$ and a place action $\hat{a}
_{\mathrm{pl}}^t$ to initiate state transition from $s^t$ to $s^{t+1}$ (\cf Eq.~\ref{eq:at_def} and Eq.~\ref{eq:state_transition}). The proposed policy generates pick-and-place actions by predicting spatial heatmaps over a discretized workspace, enabling workspace-wide reasoning while preserving spatial structure.

We implement the policy end-to-end using an encoder-decoder structured neural network $\Psi_\theta$ that predicts both pick and place spatial heatmaps. To enable a single network to perform both pick and place prediction, we introduce two conditioning inputs in addition to current and goal visual representations ($I^t$ and $I^g$): an object-conditioning map $G^t$ and a binary phase indicator map $\phi^t$. Both $G^t$ and $\phi^t$ are single-channel image-sized maps with the same spatial dimensions as the visual representations. During pick prediction, no object has yet been selected. Therefore, a zero conditioning map $G^t_\emptyset=\mathbf{0}$ and phase indicator map $\phi^t_{\mathrm{pk}}=\mathbf{0}$ are provided to the network. During place prediction, an object-specific conditioning map is constructed from the selected object and the phase indicator is set to $\phi^t_{\mathrm{pl}}=\mathbf{1}$. These conditioning signals allow the same network to distinguish between the two prediction stages while sharing the same parameters.\par
For the pick action prediction, given the current $I^t$ and the goal visual representations $I^g$, the neural network $\Psi_\theta$ predicts a pick heatmap $H^t_{\mathrm{pk}} \in \mathbb{R}^{m\times m}$, where each cell encodes the likelihood of selecting the corresponding workspace location.
\begin{equation}\label{eq6}
    H^t_{\mathrm{pk}} = \Psi_\theta(I^t, I^g, G^t_\emptyset, \phi^t_{\mathrm{pk}})
\end{equation}
A pick action location $a^t_{\mathrm{pk}}$ is subsequently extracted from $H^t_{\mathrm{pk}}$ using the action selection procedure described in Section IV.D.\par
Since the predicted pick location may not exactly coincide with an object center, it is associated with the nearest object $o_k$ in the current state:
\begin{equation}
    o_k = \argmin_{o_i \in \mathcal{O}} \lVert a^t_{\mathrm{pk}} - s^t[o_i] \rVert
\end{equation}
This association converts the spatial pick prediction into a concrete object identity, which is subsequently used to condition the place action prediction. 
Next, given the selected object $o_k$, its current location $s^t[o_k]$ is encoded into an object-specific conditioning map. Specifically, we construct a Gaussian indicator map
\begin{equation}
    G^t = \Gamma(s^t[o_k])
\end{equation}
where $\Gamma(\cdot)$ denotes the Gaussian indicator map centered at the current location of the selected object $o_k$. This indicator map explicitly identifies the selected object and provides object-aware context for the subsequent place action prediction. 
After the object-specific conditioning map is constructed, the network performs a second inference pass to predict a place heatmap $H^t_{\mathrm{pl}} \in \mathbb{R}^{m\times m}$, where each cell encodes the likelihood of placing the selected object at the corresponding workspace location.
\begin{equation}\label{eq9}
    H^t_{\mathrm{pl}}= \Psi_\theta(I^t, I^g, G^t, \phi^t_{\mathrm{pl}})
\end{equation}
A place action location $a^t_{\mathrm{pl}}$ is similarly extracted from $H^t_{\mathrm{pl}}$ using the same action selection procedure in Section IV.D.\par
Although Equations \eqref{eq6} and \eqref{eq9} are written as separate pick and place prediction passes for clarity, both predictions are generated using the same network parameters $\theta$. The distinction arises solely from the object-conditioning map and phase indicator provided as network inputs, allowing a single network to perform both pick and place prediction while sharing the same parameters. \par

\subsection{Action Refinement Strategy (ARS)}

The learned visual policy predicts probabilistic pick-and-place heatmaps over the workspace, where high-confidence predictions are not necessarily executable in cluttered non-monotone rearrangement tasks. To convert these predictions into executable actions, we propose the \textbf{A}ction \textbf{R}efinement \textbf{S}trategy (ARS), an inference-time action selection framework comprising three sequential stages: diversity-promoting candidate exploration, feasibility validation, and self-placement rejection.

Rather than committing to the highest-scoring heatmap peak, ARS first extracts the top-$K$ candidate locations. Their confidence values are normalized into a multinomial distribution, from which candidate actions are sampled to produce a prioritized evaluation order. This preserves the preference for high-confidence predictions while allowing alternative candidates to be considered when the highest-scoring prediction is infeasible.

The sampled candidate actions are then validated before execution using the Linear Motion Planner (LMP) \cite{ren2024multi}, which employs a rectangular swept-volume approximation for computationally efficient feasibility checking. Specifically, LMP verifies grasp accessibility for pick actions and rearrangement feasibility for place actions according to the constraints described in Section~III. Invalid candidates are discarded until a feasible action is found. Finally, self-placement actions are rejected by temporarily excluding the selected object and recomputing the pick and place predictions. Together, these stages transform probabilistic heatmap predictions into feasible and effective rearrangement actions without additional training or expensive online replanning.

\subsection{Progress Enforcement Strategy (PES)}

To mitigate repetitive relocation loops and improve long-horizon execution, we propose the \textbf{P}rogress \textbf{E}nforcement \textbf{S}trategy (PES), an inference-time reasoning module that performs a short-horizon rollout probe to evaluate whether the action selected by ARS is likely to sustain meaningful progress toward the goal configuration. While ARS ensures that the selected action is feasible, learning-based visual policies may still repeatedly select the same object, leading to little progress toward the target arrangement.

Starting from the current state, PES virtually executes the action proposed by ARS and rolls the policy forward for several decision steps using the same goal-conditioned policy and ARS action selection. If the rollout repeatedly selects the same object, the current action is classified as leading to progress stagnation. The selected object is then temporarily excluded from the candidate set, allowing ARS to select an alternative object for relocation. The resulting action is executed directly without performing an additional rollout probe.

Unlike ARS, which improves immediate action feasibility, PES evaluates the short-horizon consequences of candidate actions before execution to prevent repetitive relocation loops. Together, ARS and PES improve both immediate action quality and long-horizon rearrangement performance without additional training or expensive online replanning.

\subsection{Visual Adaptive Rearrangement Planner (VARP)}
\vspace{-0.4cm}
\begin{algorithm}[hbt!]
\caption{Visual Adaptive Rearrangement Planner}
\label{alg:cvrp}
\KwData{$s^0, s^g, \Psi_{\theta}, T_{\max}$}
\KwResult{$\mathcal{T}=\{a^0,a^1,\dots\}$ \Comment{Final Action Sequence}} 
$\mathcal{T} \leftarrow \emptyset$, $t\leftarrow 0$ \;
\While{$s^t\neq s^g$ \textbf{and} $t<T_{\max}$}{
$I^t \leftarrow \Phi(s^t)$, $I^g \leftarrow \Phi(s^g)$ \Comment{Render Visual States}\;
$H^t_{\mathrm{pk}} \leftarrow \Psi_\theta(I^t,I^g,G^t_{\emptyset},\mathbf{0})$ \Comment{Pick Heatmap}\;
$\hat{a}^t_{\mathrm{pk}} \leftarrow \text{ARS}_{\mathrm{pk}} (H^t_{\mathrm{pk}},s^t)$ \Comment{Action Refinement}\;
$o_k \leftarrow \argmin_{o_i\in\mathcal{O}}
\lVert \hat{a}^t_{\mathrm{pk}}-s^t[o_i]\rVert$ \Comment{Association}\;
$\hat{a}^t_{\mathrm{pk}} \leftarrow s^t[o_k]$ \Comment{Predicted Pick Action}\;
$G^t \leftarrow \Gamma(s^t[o_k])$ \Comment{Pick Object Indicator}\;
$H^t_{\mathrm{pl}} \leftarrow \Psi_\theta(I^t,I^g,G^t,\mathbf{1})$\Comment{Place Heatmap}\;
$\hat{a}^t_{\mathrm{pl}} \leftarrow \text{ARS}_{\mathrm{pl}} (H^t_{\mathrm{pl}},s^t,o_k)$\;
$\hat{a}^t \leftarrow (\hat{a}^t_{\mathrm{pk}},\hat{a}^t_{\mathrm{pl}})$ \Comment{Proposed Action}\;
$a^t \leftarrow \text{PES}(\hat{a}^t)$ \Comment{Progress Enforcement}\;
$\mathcal{T} \text{.add}(a^t$) \Comment{Append to Final Sequence}\;
$s^{t+1} \leftarrow f(s^t,a^t)$ \Comment{New State Transition}\;
$t\leftarrow t+1$ \Comment{Time Step Update}\;
}
\Return $\mathcal{T}$ if $s^t=s^g$ else $\emptyset$\;
\end{algorithm}

Algorithm \ref{alg:cvrp} summarizes the proposed Visual Adaptive Rearrangement Planner (VARP), which integrates the goal-conditioned visual policy, Action Refinement Strategy (ARS), and Progress Enforcement Strategy (PES) into a unified closed-loop framework.

Given the initial and goal states, VARP first predicts a pick heatmap using the goal-conditioned visual policy (Lines 3--4). ARS then performs candidate exploration and feasibility validation to select a manipulation object (Lines 5--8). Conditioned on the selected object, the policy predicts a place heatmap, and ARS is applied again to obtain a feasible placement location (Lines 9--11).

The resulting pick-and-place action is evaluated by PES through a short-horizon rollout probe to detect progress stagnation (Line 12). If stagnation is detected, the selected object is excluded and ARS generates an alternative action, which is accepted without further PES evaluation.

The validated action is appended to the solution sequence and executed to obtain the next state (Lines 13--15). This process repeats until the goal configuration is reached or the maximum execution horizon is exceeded (Line 2), after which VARP returns the generated action sequence $\tau=\{a^0,a^1,\ldots\}$ (Line 16).

\subsection{Expert Demonstration Generation}
The proposed visual policy is trained on demonstrations generated by the Multi-Stage Monte Carlo Tree Search (MS-MCTS) rearrangement planner \cite{ren2024multi} in confined $20\times20$ planar workspaces containing $4$ to $8$ cylindrical objects with radius 1. MS-MCTS formulates rearrangement as a 2D planning problem on the workspace plane and addresses non-monotone rearrangement by decomposing object dependencies into subgoals and jointly reasoning over object relocations and robot actions. We adopt the same 2D formulation for visual policy learning.

For each rearrangement instance, MS-MCTS generates a successful trajectory from the initial configuration $s^0$ to the goal configuration $s^g$. The trajectory is decomposed into state-action pairs $(s^t,a^t)$ and rendered into training tuples $(I^t,I^g,a^t)$. Aggregating all successful trajectories forms the demonstration dataset for supervised policy learning.

\subsection{Policy Learning}


Given the MS-MCTS demonstration dataset, each expert action $a^t=(a^t_{\mathrm{pk}},a^t_{\mathrm{pl}})$ is converted into target pick and place cells on the workspace grid. The network is trained via supervised imitation learning using the joint objective

\begin{equation}
L=\alpha L_{\mathrm{pk}}+(1-\alpha) L_{\mathrm{pl}}
\end{equation}
where $L_{\mathrm{pk}}$ and $L_{\mathrm{pl}}$ are cross-entropy losses for the pick and place predictions, while $\alpha$ is the hyperparameter for adjusting loss weights. 

VARP uses an encoder-decoder neural network to learn dense spatial action distributions, preserving object spatial relationships rather than directly regressing object coordinates. This representation naturally captures multiple plausible manipulation actions, enabling the downstream ARS and PES modules to refine actions and enforce execution progress at inference time.

\subsection{Implementation Details}
The goal-conditioned visual policy uses a U-Net-based \cite{ronneberger2015u} encoder-decoder architecture with 4.2M trainable parameters. The network receives an 8-channel input consisting of the current and goal RGB visual state representations, a pick-indicator channel, and a phase channel. A shared network predicts both pick and place heatmaps over the discretized workspace.

The policy is trained by supervised imitation learning on MS-MCTS demonstrations. Expert pick and place actions are discretized onto the $20\times20$ workspace grid and used as supervision targets. Training uses AdamW \cite{loshchilov2017decoupled} with a learning rate of $3\times10^{-4}$, batch size 128, and a combined pick-place cross-entropy loss with label smoothing \cite{szegedy2016rethinking} ($\alpha=0.5$). Teacher forcing is applied during place prediction.

During inference, ARS performs top-$5$ candidate exploration, feasibility validation, and self-placement rejection, while PES uses a three-step rollout probe to detect progress stagnation. All experiments use a maximum execution budget of $5N$ actions, where $N$ is the number of objects.

The final dataset contains 100,012 demonstrations (1M state-action transitions), requiring 34.7 hours to generate on an Intel Xeon Gold 6242R CPU. Training on a single NVIDIA Tesla V100 GPU required 6.5 hours for 10 epochs.

\section{EXPERIMENTS}
\subsection{Experiment Setup}

We evaluate VARP on non-monotone object rearrangement tasks in a synthetic discretized confined workspace of size $20\times20$ units. Each instance contains $N$ cylindrical objects with radius of 1 unit, where $N\in{\{4,5,6,7,8\}}$, representing increasing levels of task difficulty. All methods use the same Linear Motion Planner (LMP) \cite{ren2024multi} for motion generation to ensure a fair comparison of high-level planning performance. Initial and goal configurations are sampled to be collision-free. Due to the confined and cluttered workspace, many instances require intermediate object relocations before all objects can be placed at their target locations.

We first compare VARP against three representative baselines: MS-MCTS \cite{ren2024multi}, a search-based expert planner; PERTS (CIRS) \cite{wang2022efficient}, a perturbation-based theoretically complete rearrangement planner; and the Transporter Network \cite{zeng2021transporter}, a learning-based manipulation policy, on a randomly generated standard rearrangement benchmark. We then evaluate VARP on a disturbance recovery benchmark against the expert planner (MS-MCTS), followed by ablation studies, out-of-distribution generalization, robustness, and real-world evaluations.

\noindent\textbf{Evaluation Metrics:} The following metrics are used for quantitative evaluation:

\begin{itemize}
\item \textbf{Success Rate (SR) (\%)}: The percentage of rearrangement instances that reach the goal configuration within the evaluation budget, where search-based methods are given a 60 s planning budget and learning-based methods are allowed up to 5N actions where N is the number of objects in the scene.

\item \textbf{Planning Time (PT) (s)}: The average end-to-end computation time required to generate a rearrangement solution before execution, measured in seconds.

\item \textbf{Number of Steps (NS)}: The total number of pick-and-place actions executed to complete the task.

\item \textbf{Object Displacement Distance (DD)}: The cumulative distance traveled by all manipulated objects throughout the rearrangement process.

\end{itemize}

\subsection{Standard Rearrangement Benchmark}

\begin{table}[t]
\centering
\footnotesize
\setlength{\tabcolsep}{3pt}
\caption{Performance on the standard rearrangement benchmark containing 2,000 evaluation instances. VARP outperforms all baselines in success rate while maintaining sub-second planning time.}
\label{tab:final2000_metrics}
\begin{tabular}{lcccc}
\toprule
\textbf{Method} & \textbf{SR $\uparrow$}  & \textbf{PT $\downarrow$} & \textbf{NS $\downarrow$} & \textbf{DD $\downarrow$} \\
\midrule
\textbf{VARP (Ours)}
& \textbf{98.25}
& 0.40 $\pm$ 0.26
& 13.59 $\pm$ 7.08
& 88.99 $\pm$ 39.04 \\

MS-MCTS
& 89.50 
& 8.67 $\pm$ 11.30
& 9.46 $\pm$ 3.51
& 76.06 $\pm$ 28.60 \\

PERTS
& 68.30 
& 2.20 $\pm$ 6.81
& 40.27 $\pm$ 49.58
& 351.12 $\pm$ 406.53 \\

Transporter
& 21.85
& 0.06 $\pm$ 0.03
& 7.31 $\pm$ 2.84 
& 56.45 $\pm$ 19.32 \\
\hline
\end{tabular}
\vspace{-0.3cm}
\end{table}

The standard benchmark consists of 2000 rearrangement instances generated by randomly sampling collision-free initial and goal configurations in the confined 20×20 workspace with only the front opening, consisting of 400 instances for each object count from 4 to 8. The constrained workspace frequently requires intermediate object relocations during planning.

Table \ref{tab:final2000_metrics} summarizes the results on the 2,000-instance benchmark. VARP achieves the highest success rate among all methods while requiring only 0.40 s of planning time on average, providing over a 20× speedup compared with the expert. These results demonstrate that VARP effectively handles the long-horizon object dependencies required for confined rearrangement. Although MS-MCTS typically produces slightly shorter rearrangement sequences through explicit online search, it incurs substantially higher computational cost and solves fewer instances. In contrast, VARP prioritizes robust task completion with only a modest increase in relocation actions. PERTS struggles with the complex dependencies and intermediate relocations of confined rearrangement, while the Transporter Network performs substantially worse due to its limited long-horizon reasoning capability.

\subsection{Disturbance Recovery Benchmark}
\begin{table*}[t]
\centering
\caption{Performance on the disturbance recovery benchmark. VARP achieves higher overall success rates and substantially lower planning times than the expert planner (Exp) under execution-time disturbances.}
\label{tab:reactive_benchmark}
\begin{tabular}{ccccc}
\toprule
\textbf{$K$} &
\textbf{SR $\uparrow$ (Ours / Exp)} &
\textbf{PT $\downarrow$ (Ours / Exp)} &
\textbf{NS $\downarrow$ (Ours / Exp)} &
\textbf{DD $\downarrow$ (Ours / Exp)} \\
\midrule

2 &
\textbf{100.0} / 98.0 &
\textbf{0.19 $\pm$ 0.07} / 2.90 $\pm$ 2.35 &
7.26 $\pm$ 2.06 / \textbf{6.90 $\pm$ 1.53} &
\textbf{52.24 $\pm$ 11.18} / 52.32 $\pm$ 11.33 \\

3 &
96.0 / \textbf{98.0} &
\textbf{0.25 $\pm$ 0.12} / 7.33 $\pm$ 10.44 &
9.10 $\pm$ 2.60 / \textbf{8.39 $\pm$ 1.87} &
63.65 $\pm$ 18.58 / \textbf{61.88 $\pm$ 15.41} \\

4 &
\textbf{100.0} / 90.0 &
\textbf{0.36 $\pm$ 0.13} / 13.71 $\pm$ 13.22 &
12.68 $\pm$ 3.50 / \textbf{11.27 $\pm$ 2.65} &
86.05 $\pm$ 25.11 / \textbf{83.02 $\pm$ 23.94} \\

5 &
\textbf{94.0} / 78.0 &
\textbf{0.54 $\pm$ 0.22} / 17.58 $\pm$ 12.17 &
18.38 $\pm$ 5.78 / \textbf{13.82 $\pm$ 2.37} &
117.68 $\pm$ 27.64 / \textbf{105.83 $\pm$ 21.71} \\

\midrule

ALL &
\textbf{97.5} / 91.0 &
\textbf{0.33 $\pm$ 0.20} / 9.91 $\pm$ 11.70 &
11.78 $\pm$ 5.63 / \textbf{9.86 $\pm$ 3.36} &
79.49 $\pm$ 32.83 / \textbf{73.95 $\pm$ 27.28} \\

\bottomrule
\end{tabular}
\end{table*}

To evaluate the adaptive capability of VARP, we introduce a reactive disturbance benchmark. Both VARP and the expert planner first execute three identical relocation actions from a shared initial state. A disturbance is then introduced by randomly relocating 2 to 5 objects to collision-free positions, with 50 instances per disturbance level (200 tasks in total). Following the disturbance, the expert planner is allowed a single replanning step, while VARP continues execution using closed-loop visual feedback.

Table \ref{tab:reactive_benchmark} reports the results, where $k$ denotes the number of disturbed objects. VARP achieves a $97.5\%$ success rate, outperforming the expert by 6.5\% while reducing the average planning time from 9.91s to 0.33s. The improvement is statistically significant (paired McNemar's test \cite{mcnemar1947note}, $\chi^2=8.47$, $p=3.6\times10^{-3}$). As the disturbance level increases, VARP consistently widens its performance advantage, demonstrating strong robustness to execution-time disturbances without repeated replanning.

\subsection{Ablation Studies}
\begin{table*}[t]
\centering
\caption{Ablation study of the inference-time reasoning modules. Both ARS and PES consistently improve the planner's performance, while increasing the amount of expert demonstrations further enhances the success rate. }
\label{tab:ablation}
\begin{tabular}{lcccc}
\toprule
\textbf{Variant} & \textbf{SR $\uparrow$ (N=4 $|$ 5 $|$ 6 $|$ 7 $|$ 8 $|$ Avg.)} & \textbf{PT $\downarrow$} & \textbf{NS $\downarrow$ } & \textbf{DD $\downarrow$} \\
\midrule
VARP (ARS + PES)                &  100.0 $|$ 99.20 $|$ 98.80 $|$ 97.80 $|$ 95.50 $|$  98.25 & 0.40 $\pm$ 0.26 & 13.59 $\pm$ 7.08 & 88.99 $\pm$ 39.04 \\
VARP w/o PES (ARS only)         &  99.50  $|$ 98.20 $|$ 97.50 $|$ 96.20 $|$ 91.50 $|$ 96.60          & 0.21 $\pm$ 0.12 & 13.58 $\pm$ 7.01 & 88.92 $\pm$ 39.26 \\
VARP w/o (ARS \& PES)   & 70.80 $|$ 51.50 $|$ 36.20 $|$ 17.20 $|$ 08.00 $|$ 36.75          & 0.09 $\pm$ 0.04 & 8.30 $\pm$ 3.30  & 62.87 $\pm$ 22.79 \\
VARP w/ 40k training Data     &  99.80  $|$ 99.80 $|$ 98.50 $|$ 97.50 $|$ 91.80 $|$ 97.45          & 0.40 $\pm$ 0.25 & 13.79 $\pm$ 7.09 & 89.14 $\pm$ 38.43 \\
\bottomrule
\end{tabular}
\vspace{-0.4cm}
\end{table*}

Table~\ref{tab:ablation} evaluates the contributions of ARS, PES, and training data scale on the same standard rearrangement benchmark mentioned in Section V-B. Removing both ARS and PES reduces the success rate from 98.25\% to 36.75\%, indicating that the learned policy alone is insufficient for reliable execution in cluttered non-monotone rearrangement tasks. Adding ARS increases the success rate to 96.60\% by enforcing feasibility-aware action selection and eliminating most invalid or self-defeating actions. PES further improves performance to 98.25\%, with the largest gains observed on dense scenes, \textit{e.g.}, 91.5\% to 95.5\% on 8-object instances, demonstrating its effectiveness in mitigating residual progress stagnation behaviors. Finally, increasing the training set from 40k to 100k demonstrations improves overall success from 97.45\% to 98.25\%, with the most noticeable benefit occurring on the most challenging high-cluttered instances.

\subsection{Generalization and Robustness Evaluation}
Beyond benchmark performance, we further evaluate the generalization and robustness of VARP. Specifically, we investigate the ability of the learned policy to handle problem settings that differ from the training distribution and assess its sensitivity to visual perturbations in the environment. These experiments provide additional insight into the scalability and practical reliability of the proposed framework.

\begin{table}[t]
\caption{Robustness and out-of-distribution generalization results. VARP remains robust under continuous disturbances and generalizes to unseen object counts beyond the training distribution.}
\label{tab:robust_ood}
\centering
\footnotesize
\setlength{\tabcolsep}{4pt}

\begin{tabular}{cccc}
\toprule
\multicolumn{4}{c}{\textbf{(a) Continuous Disturbance Stress Test}} \\
\midrule
\textbf{$N$} & \textbf{SR $\uparrow$ (Ours / Exp)} & \textbf{PT $\downarrow$ (Ours / Exp)}  & \textbf{NS $\downarrow$ (Ours / Exp)}\\
\midrule
4 & \textbf{100.0 / 100.0} & \textbf{0.20} / 3.16  & 6.75 / \textbf{6.08}\\
5 & \textbf{100.0} / 97.5 & \textbf{0.37} / 7.55 &  11.05 / \textbf{10.15}\\
6 & \textbf{97.5} / 67.5 & \textbf{0.51} / 23.46 &  \textbf{15.92} / 19.70\\
7 & \textbf{97.5} / 42.5 & \textbf{0.69} / 45.96 &  \textbf{21.00} / 26.24\\
8 & \textbf{77.5} / 5.0 & \textbf{1.22} / 139.91 &  \textbf{34.26} / 60.00\\
\midrule
ALL & \textbf{94.5} / 62.5 & \textbf{0.57} / 16.93 & 17.01 / \textbf{13.90}\\
\midrule
\multicolumn{4}{c}{\textbf{(b) Out-Of-Distribution Generalization}} \\
\midrule
\textbf{$N$} & \textbf{SR} & \textbf{PT}  & \textbf{NS} \\
\midrule
2 & 100.0 & 0.06  & 2.51 \\
3 & 100.0 & 0.10 & 4.16 \\
9 & 91.0 & 0.92  & 28.40 \\
10 & 64.0 & 1.16 & 35.48 \\
\midrule
ALL & 88.8 & 0.49 & 15.55\\
\bottomrule
\end{tabular}
\vspace{-0.3cm}
\end{table}

\subsubsection{Continuous Disturbance Stress Test}

To evaluate robustness under persistent environmental changes, we introduce a continuous disturbance benchmark in which two objects are randomly relocated every two relocation actions throughout execution. This setting requires the planner to repeatedly adapt while continuing to progress toward the target arrangement.

Table \ref{tab:robust_ood}-(a) reports the results. VARP achieves an overall success rate of 94.5\%, outperforming the expert planner by 32\% while requiring substantially less planning time. The performance gap further increases with task complexity, demonstrating that VARP remains robust under sustained execution-time disturbances.

\subsubsection{Out-of-Distribution Generalization}

To evaluate the scalability and generalization capability of VARP, we train the policy on rearrangement tasks containing 4 to 8 objects and evaluate it on unseen scenes containing 2, 3, 9, and 10 objects without additional retraining or finetuning.

As shown in Table \ref{tab:robust_ood}-(b), VARP achieves a 100\% success rate on 2 and 3 objects scenes and maintains success rates of 91\% and 64\% on 9 and 10 objects scenes, respectively. These results demonstrate that VARP generalizes effectively beyond the training distribution to both simpler and more complex rearrangement tasks.

\subsection{Real-World Experiments}

We validate the proposed VARP on a Trossen ViperX-300S 6-DOF robot arm in a confined workspace measuring $52~\mathrm{cm}\times40~\mathrm{cm}$ with cylindrical objects of diameter $6~\mathrm{cm}$. The robot accesses the workspace exclusively through the front opening, matching the manipulation protocol used in simulation. VARP is trained from expert demonstrations generated using the real robot workspace geometry with 4 and 5 objects.

We then evaluate VARP on 11 real-world rearrangement instances spanning three scenarios: (1) standard in-distribution tasks with 4 and 5 objects, (2) out-of-distribution tasks containing 3 and 6 objects, and (3) disturbance recovery tasks, where objects are manually displaced during execution. For each instance, the robot is initialized to a randomly generated start configuration and executes VARP in a closed-loop manner until the target arrangement is reached.

VARP successfully solves all 11 real-world instances. One challenging four-object in-place $90^\circ$ rotation example is shown in Fig.~\ref{fig:real_exp1}. The policy performs real-time inference, requiring only $0.0187$ s per decision and $0.136$ s per instance on average. Despite being trained only on 4- and 5-object demonstrations, VARP generalizes to both smaller (3) and larger (6) object sets and consistently recovers from execution-time disturbances. These results demonstrate effective sim-to-real transfer with fast closed-loop execution, strong generalization, and robust disturbance recovery. Additional examples are provided in the supplementary video.

\section{CONCLUSION}

This paper presented VARP, a visual adaptive rearrangement planner for non-monotone object rearrangement planning in confined environments. By combining a goal-conditioned visual policy with the proposed Action Refinement Strategy (ARS) and Progress Enforcement Strategy (PES), VARP transforms and refines raw visual policy predictions into feasible and progress-oriented rearrangement actions without additional training. Extensive simulation and real-world experiments demonstrate that VARP achieves high success rates while substantially reducing planning time compared with the state-of-the-art baselines. Furthermore, VARP exhibits strong robustness to execution-time disturbances and generalizes beyond the training distribution to unseen object counts. Future work will extend VARP to more complex manipulation settings, including multi-arm planning problems and in unstructured environments.

\bibliographystyle{IEEEtran}
\bibliography{root}

@inproceedings{ren2024multi,
  title={Multi-stage monte carlo tree search for non-monotone object rearrangement planning in narrow confined environments},
  author={Ren, Hanwen and Qureshi, Ahmed H},
  booktitle={2024 IEEE/RSJ International Conference on Intelligent Robots and Systems (IROS)},
  pages={12078--12085},
  year={2024},
  organization={IEEE}
}

@inproceedings{ronneberger2015u,
  title={U-net: Convolutional networks for biomedical image segmentation},
  author={Ronneberger, Olaf and Fischer, Philipp and Brox, Thomas},
  booktitle={International Conference on Medical image computing and computer-assisted intervention},
  pages={234--241},
  year={2015},
  organization={Springer}
}

@article{loshchilov2017decoupled,
  title={Decoupled weight decay regularization},
  author={Loshchilov, Ilya and Hutter, Frank},
  journal={arXiv preprint arXiv:1711.05101},
  year={2017}
}

@inproceedings{szegedy2016rethinking,
  title={Rethinking the inception architecture for computer vision},
  author={Szegedy, Christian and Vanhoucke, Vincent and Ioffe, Sergey and Shlens, Jon and Wojna, Zbigniew},
  booktitle={Proceedings of the IEEE conference on computer vision and pattern recognition},
  pages={2818--2826},
  year={2016}
}

@inproceedings{zeng2021transporter,
  title={Transporter networks: Rearranging the visual world for robotic manipulation},
  author={Zeng, Andy and Florence, Pete and Tompson, Jonathan and Welker, Stefan and Chien, Jonathan and Attarian, Maria and Armstrong, Travis and Krasin, Ivan and Duong, Dan and Sindhwani, Vikas and others},
  booktitle={Conference on Robot Learning},
  pages={726--747},
  year={2021},
  organization={PMLR}
}

@inproceedings{wang2022efficient,
  title={Efficient and high-quality prehensile rearrangement in cluttered and confined spaces},
  author={Wang, Rui and Miao, Yinglong and Bekris, Kostas E},
  booktitle={2022 International Conference on Robotics and Automation (ICRA)},
  pages={1968--1975},
  year={2022},
  organization={IEEE}
}

@inproceedings{shridhar2021cliport,
  title     = {{CLIPort}: What and Where Pathways for Robotic Manipulation},
  author    = {Shridhar, Mohit and Manuelli, Lucas and Fox, Dieter},
  booktitle = {Conference on Robot Learning (CoRL)},
  year      = {2021}
}

@inproceedings{shridhar2022perceiveractor,
  title     = {Perceiver-Actor: A Multi-Task Transformer for Robotic Manipulation},
  author    = {Shridhar, Mohit and Manuelli, Lucas and Fox, Dieter},
  booktitle = {Conference on Robot Learning (CoRL)},
  year      = {2022}
}

@inproceedings{jiang2023vima,
  title     = {{VIMA}: General Robot Manipulation with Multimodal Prompts},
  author    = {Jiang, Yunfan and Gupta, Agrim and Zhang, Zichen and Wang, Guanzhi and
               Dou, Yongqiang and Chen, Yanjun and Fei-Fei, Li and Anandkumar, Anima and
               Zhu, Yuke and Fan, Linxi},
  booktitle = {International Conference on Machine Learning (ICML)},
  year      = {2023}
}

@inproceedings{goyal2022ifor,
  title     = {{IFOR}: Iterative Flow Minimization for Robotic Object Rearrangement},
  author    = {Goyal, Ankit and Mousavian, Arsalan and Paxton, Chris and Chao, Yu-Wei and
               Okorn, Brian and Deng, Jia and Fox, Dieter},
  booktitle = {IEEE/CVF Conference on Computer Vision and Pattern Recognition (CVPR)},
  year      = {2022}
}

@inproceedings{liu2022structformer,
  title     = {{StructFormer}: Learning Spatial Structure for Language-Guided Semantic
               Rearrangement of Novel Objects},
  author    = {Liu, Weiyu and Paxton, Chris and Hermans, Tucker and Fox, Dieter},
  booktitle = {IEEE International Conference on Robotics and Automation (ICRA)},
  year      = {2022}
}

@inproceedings{qureshi2021nerp,
  title     = {{NeRP}: Neural Rearrangement Planning for Unknown Objects},
  author    = {Qureshi, Ahmed H. and Mousavian, Arsalan and Paxton, Chris and
               Yip, Michael C. and Fox, Dieter},
  booktitle = {Robotics: Science and Systems (RSS)},
  year      = {2021}
}

@article{stilman2008planning,
  title   = {Planning Among Movable Obstacles with Artificial Constraints},
  author  = {Stilman, Mike and Kuffner, James J.},
  journal = {The International Journal of Robotics Research (IJRR)},
  volume  = {27},
  number  = {11--12},
  pages   = {1295--1307},
  year    = {2008}
}

@inproceedings{king2016rearrangement,
  title     = {Rearrangement Planning Using Object-Centric and Robot-Centric Action Spaces},
  author    = {King, Jennifer E. and Cognetti, Marco and Srinivasa, Siddhartha S.},
  booktitle = {IEEE International Conference on Robotics and Automation (ICRA)},
  year      = {2016}
}

@inproceedings{krontiris2015dealing,
  title     = {Dealing with Difficult Instances of Object Rearrangement},
  author    = {Krontiris, Athanasios and Bekris, Kostas E.},
  booktitle = {Robotics: Science and Systems (RSS)},
  year      = {2015}
}

@article{labbe2020monte,
  title   = {Monte-Carlo Tree Search for Efficient Visually Guided Rearrangement Planning},
  author  = {Labb{\'e}, Yann and Zagoruyko, Sergey and Kalevatykh, Igor and Laptev, Ivan and
             Carpentier, Justin and Aubry, Mathieu and Sivic, Josef},
  journal = {IEEE Robotics and Automation Letters (RA-L)},
  volume  = {5},
  number  = {2},
  pages   = {3715--3722},
  year    = {2020}
}

@article{ren2026multi,
  title={Multi-Agent Monte Carlo Tree Search for Makespan-Efficient Object Rearrangement in Cluttered Spaces},
  author={Ren, Hanwen and Kim, Junyong and Tharmasanthiran, Aathman and Qureshi, Ahmed H},
  journal={arXiv preprint arXiv:2602.02411},
  year={2026}
}

@inproceedings{chen2024language,
  title={Language-guided Active Sensing of Confined, Cluttered Environments via Object Rearrangement Planning},
  author={Chen, Weihan and Ren, Hanwen and Qureshi, Ahmed H},
  booktitle={2024 IEEE International Conference on Robotics and Automation (ICRA)},
  pages={11224--11230},
  year={2024},
  organization={IEEE}
}

@inproceedings{kim2025integrating,
  title={Integrating active sensing and rearrangement planning for efficient object retrieval from unknown, confined, cluttered environments},
  author={Kim, Junyong and Ren, Hanwen and Qureshi, Ahmed H},
  booktitle={2025 IEEE International Conference on Robotics and Automation (ICRA)},
  pages={12572--12578},
  year={2025},
  organization={IEEE}
}

@inproceedings{ren2024neural,
  title={Neural rearrangement planning for object retrieval from confined spaces perceivable by robot’s in-hand rgb-d sensor},
  author={Ren, Hanwen and Qureshi, Ahmed H},
  booktitle={2024 IEEE International Conference on Robotics and Automation (ICRA)},
  pages={15388--15394},
  year={2024},
  organization={IEEE}
}

@inproceedings{liu2021deep,
  title={Deep structured reactive planning},
  author={Liu, Jerry and Zeng, Wenyuan and Urtasun, Raquel and Yumer, Ersin},
  booktitle={2021 IEEE International Conference on Robotics and Automation (ICRA)},
  pages={4897--4904},
  year={2021},
  organization={IEEE}
}

@inproceedings{bowen2014closed,
  title={Closed-loop global motion planning for reactive execution of learned tasks},
  author={Bowen, Chris and Alterovitz, Ron},
  booktitle={2014 IEEE/RSJ International Conference on Intelligent Robots and Systems},
  pages={1754--1760},
  year={2014},
  organization={IEEE}
}

@article{pan2022algorithms,
  title={Algorithms and systems for manipulating multiple objects},
  author={Pan, Zherong and Zeng, Andy and Li, Yunzhu and Yu, Jingjin and Hauser, Kris},
  journal={IEEE Transactions on Robotics},
  volume={39},
  number={1},
  pages={2--20},
  year={2022},
  publisher={IEEE}
}

@article{mcnemar1947note,
  title={Note on the sampling error of the difference between correlated proportions or percentages},
  author={McNemar, Quinn},
  journal={Psychometrika},
  volume={12},
  number={2},
  pages={153--157},
  year={1947},
  publisher={Springer-Verlag}
}

@inproceedings{wilfong1988motion,
  title={Motion planning in the presence of movable obstacles},
  author={Wilfong, Gordon},
  booktitle={Proceedings of the fourth annual symposium on Computational geometry},
  pages={279--288},
  year={1988}
}

@article{kappler2018real,
  title={Real-time perception meets reactive motion generation},
  author={Kappler, Daniel and Meier, Franziska and Issac, Jan and Mainprice, Jim and Cifuentes, Cristina Garcia and W{\"u}thrich, Manuel and Berenz, Vincent and Schaal, Stefan and Ratliff, Nathan and Bohg, Jeannette},
  journal={IEEE Robotics and Automation Letters},
  volume={3},
  number={3},
  pages={1864--1871},
  year={2018},
  publisher={IEEE}
}

@article{han2017high,
  title={High-quality tabletop rearrangement with overhand grasps: Hardness results and fast methods},
  author={Han, Shuai D and Stiffler, Nicholas M and Krontiris, Athansios and Bekris, Kostas E and Yu, Jingjin},
  journal={arXiv preprint arXiv:1705.09180},
  year={2017}
}

@inproceedings{yuan2018rearrangement,
  title={Rearrangement with nonprehensile manipulation using deep reinforcement learning},
  author={Yuan, Weihao and Stork, Johannes A and Kragic, Danica and Wang, Michael Y and Hang, Kaiyu},
  booktitle={2018 IEEE International Conference on Robotics and Automation (ICRA)},
  pages={270--277},
  year={2018},
  organization={IEEE}
}

@inproceedings{tang2023selective,
  title={Selective object rearrangement in clutter},
  author={Tang, Bingjie and Sukhatme, Gaurav S},
  booktitle={Conference on Robot Learning},
  pages={1001--1010},
  year={2023},
  organization={PMLR}
}

\end{document}